\documentclass{article}
\usepackage{iclr2021_conference,times}

\usepackage{amsmath,amsfonts,bm}

\def\eqref#1{equation~\ref{#1}}

\def\1{\bm{1}}

\def\eps{{\epsilon}}

\DeclareMathAlphabet{\mathsfit}{\encodingdefault}{\sfdefault}{m}{sl}
\SetMathAlphabet{\mathsfit}{bold}{\encodingdefault}{\sfdefault}{bx}{n}

\usepackage{hyperref}
\usepackage{url}

\usepackage{graphicx}

\usepackage{subcaption}
\usepackage{wrapfig}
\usepackage[bottom]{footmisc}

\newcommand{\policy}{\pi}

\title{Improving Exploration in Policy Gradient Search: Application to Symbolic Optimization}

\author{Mikel Landajuela$^*$, Brenden K. Petersen\thanks{Equal contribution. $^\dagger$Corresponding author: \texttt{bp@llnl.gov}.}$^{\hphantom{*}\dagger}$, Soo K. Kim, Claudio P. Santiago, \\
\textbf{Ruben Glatt, T. Nathan Mundhenk, Jacob F. Pettit, \& Daniel M. Faissol} \\
Lawrence Livermore National Laboratory \\
Livermore, CA 94550, USA \\
}

\iclrfinalcopy
\begin{document}

\maketitle

\begin{abstract}

Many machine learning strategies designed to automate mathematical tasks leverage neural networks to search large combinatorial spaces of mathematical symbols.
In contrast to traditional evolutionary approaches, using a neural network at the core of the search allows learning higher-level symbolic patterns, providing an informed direction to guide the search.
When no labeled data is available, such networks can still be trained using reinforcement learning.
However, we demonstrate that this approach can suffer from an \textit{early commitment} phenomenon and from \textit{initialization bias}, both of which limit exploration.
We present two exploration methods to tackle these issues, building upon ideas of entropy regularization and distribution initialization.
We show that these techniques can improve the performance, increase sample efficiency, and lower the complexity of solutions for the task of symbolic regression.

\end{abstract}

\section{Introduction}

\begin{figure}[b]
\begin{center}
\includegraphics[trim=0 5 0 35, clip, width=0.8\textwidth]{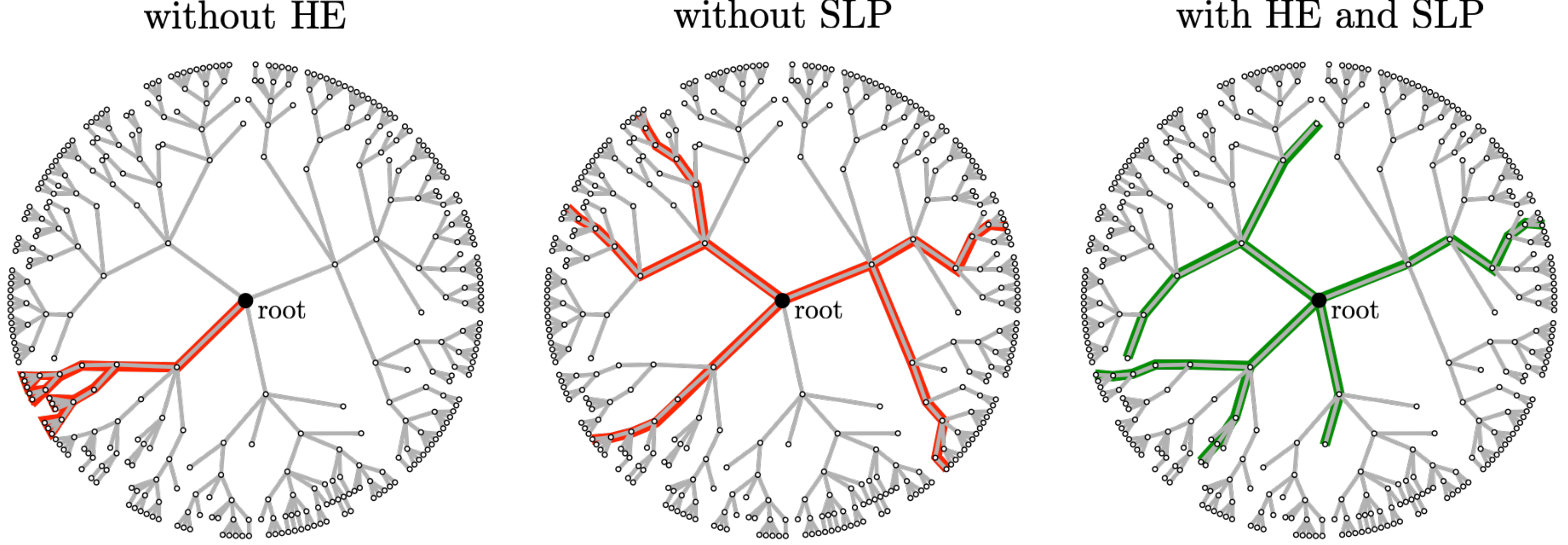}
\end{center}
\caption{
Schematic view of a symbolic search space. Any path from root to a leaf node represents a mathematical expression.
\textbf{Left.}
Without hierarchical entropy, the search tends to choose paths with the same initial branches.
\textbf{Center.}
Without the soft length prior, the search tends to reach the deepest levels of the tree, i.e. it avoids shorter expressions.
\textbf{Right.}
With both exploration techniques, the search explores various initial branches and expression lengths.
}
\label{fig:expression_space}
\end{figure}

The application of machine learning to symbolic optimization (SO) tasks such as symbolic regression (SR), automatic equation solving, 
or program synthesis involves combinatorial search spaces that are vast and complex.
In such tasks, the goal is to find a sequence of actions (i.e. symbols) that, upon semantic interpretation, optimizes a desired criteria.
For these tasks, uninformed search can suffer from prohibitively large time and memory requirements.
Thus, researchers have turned to heuristic or local search strategies, and much attention has been directed to genetic algorithms \citep{koza1992genetic, schmidt2009distilling, back2018evolutionary}.
However, the effectiveness of genetic algorithms for these tasks is still debatable \citep{russell2002artificial},
as they are known to suffer from premature convergence \citep{lu2021incorporating}, especially when the optimization landscape suffers from non-locality \citep{rothlauf2006locality}.
Approaches based on Monte Carlo Tree Search (MCTS) have also been proposed \citep{li2019neuralguided}.
However, efficient storage of the tree structure can become prohibitively large for spaces with high depth and high branching factor \citep{anthony2019policy}.

Recently, a new paradigm has emerged to use deep learning to guide search without the need to store a search tree or have access to labeled data. 
These approaches are based on deep reinforcement learning, using an autoregressive recurrent neural network (RNN) optimized via policy gradients.
Example applications range from automated machine learning (AutoML) to other SO tasks, including
neural architecture search \citep{zoph2016neural},
standard combinatorial optimization problems \citep{bello2016neural},
neural optimizer search \citep{bello2017neural},
searching for activation functions \citep{ramachandran2017searching},
program synthesis \citep{liang2018memory},
molecular optimization \citep{popova2019molecularrnn},
SR \citep{petersen2019deep},
and reinforcement learning
\citep{landajuela2021discovering}.
Software frameworks for SO include Tree-Based Pipeline Optimization Tool \citep{olson2016tpot}, PyGlove \citep{peng2021pyglove}, and Deep Symbolic Optimization\footnote{\url{https://www.github.com/brendenpetersen/deep-symbolic-optimization}}.

Given the large search space, intelligent and guided exploration is critical.
Leveraging the deep symbolic regression algorithm by \citet{petersen2019deep}, we demonstrate that standard approaches to exploration (e.g. standard entropy regularizer) can still lead to premature convergence, and that naive weight initialization can result in large initial biases of the distribution over sequence lengths.
To address these issues, we introduce two novel exploration techniques:
(1) a \textit{hierarchical entropy regularizer} to avoid premature convergence on the first few tokens
of newly generated sequences,
and (2) a \textit{soft length prior} to promote exploration over sequence lengths.
We demonstrate that each of these techniques improves both performance and sample efficiency in benchmark SR tasks.

\section{Related Work}

\textbf{Entropy exploration.}
\citet{haarnoja2018soft} demonstrate remarkable success using an entropy regularizer to aid exploration in benchmark control tasks.
However, when applied to symbolic search, this approach disproportionately encourages exploration in later search steps.
Emphasis of exploration in early steps of the tree search is considered in \citet{anthony2019policy} using a bandit rule.
However, only the first action of a trajectory is treated differently, with subsequent actions sampled normally.
In contrast, our hierarchical entropy regularizer promotes exploration across all tokens based on their position within the search tree.

\textbf{Priors.}
The use of priors is an \textit{informed search strategy} that uses knowledge about the problem to guide search and find solutions more efficiently \citep{russell2002artificial}.
Constraints have been used in policy gradient search applications:
in generating molecular structures, \citet{popova2019molecularrnn} use valency constraints to follow rules of organic chemistry;
in generating mathematical expressions, \citet{petersen2019deep} include various constraints to prevent unrealistic expressions (e.g. nested trigonometric operators) or redundancies (e.g. multiplying two constants together).
These constraints are based on domain-specific knowledge; here, in contrast, our soft length prior helps guide the search in a data-driven manner.
In other neural-guided search, \cite{li2019neuralguided} identify asymptotic constraints of leading polynomial powers and use those constraints to guide Monte Carlo tree search.

\section{Policy Gradient Search in Symbolic Spaces}
An example symbolic search space is illustrated in Fig.~\ref{fig:expression_space}.
To navigate this space, we consider \textit{policy gradient search} methods such as those discussed above.
These works employ a \textit{search policy} $\policy(\tau; \theta)$ with policy parameters $\theta$ to generate candidate solutions $\tau$.
Each candidate comprises a variable-length sequence of $T$  ``tokens'' $\tau_i$ from a library $\mathcal{L}$, such that $\tau = [ \tau_1, \dots, \tau_T]$.
For example, $\tau$ may represent a neural network architecture \citep{zoph2016neural} or a mathematical expression \citep{petersen2019deep}.
In the latter case, the library comprises binary operators (e.g. $+$ or $\div$), unary operators (e.g. $\sin$ or $\log$), and terminals (input variables or constants).
Candidates are evaluated using a black-box reward function $R(\tau) \in \mathbb{R}$.
The policy itself commonly assumes a RNN architecture.
During sampling, the RNN emits logits
$\mathbb{\psi}_{i} \in \mathbb{R}^{|\mathcal{L}|}$, 
which parameterize a categorical distribution from which the $i$th token $\tau_i$ is sampled.
Thus, the likelihood over tokens in the library at the $i$th position in the sequence is given by $\policy(\cdot | \tau_{1:(i-1)}; \theta) = \textrm{softmax}(\psi_{i})$.
Note that token $\tau_i$ is conditionally independent given the previous tokens $\tau_{1:(i-1)}$; this sampling procedure is known as \textit{autoregressive} sampling.
The search space can effectively be  pruned by penalizing \textit{in situ} the emission $\psi_{i}$ to avoid undesirable symbolic patterns (see section~\ref{sect:SLP} below).

The policy is typically trained to maximize expected reward using policy gradients \citep{williams1992simple}.
Specifically, we collect a batch of $M$ candidate solutions (or trajectories in the search space) $\{\tau^{(j)}\}_{j=1}^{M}$ sampled from the current RNN and update the parameters according to:
\begin{align}
\theta \leftarrow \theta + \alpha
\frac{1}{M} \sum_{j=1}^{ M}
\left(
 \sum_{i=1}^{|\tau^{(j)}|} \left( \left( R(\tau^{(j)}) - b \right) \nabla_\theta \log  \pi (\tau_i^{(j)} | \tau_{1:(i-1)}^{(j)}; \theta)  \right) 
+ \eta \nabla_\theta  \mathcal{R}_H (\tau^{(j)})
 \right),
\label{eqn:update-rule}
\end{align}
where $\alpha$ is the learning rate, $b$ is a control variate or ``baseline'', $\mathcal{R}_H$ is an entropy regularizer and $\eta$ is the entropy weight (see section~\ref{sect:HE} below).
We use the \textit{risk-seeking policy gradient} proposed in \citet{petersen2019deep} to promote best-case over expected performance.
This method uses $b = R_\eps$, where $R_\eps$ is the empirical $(1 -\eps)$-quantile of $\{R(\tau^{(j)})\}_{j=1}^{M}$, 
and filters the batch to contain only the top $\eps$ fraction of samples.
The training procedure presented in (\ref{eqn:update-rule}) encodes knowledge about the most promising paths in the search space in the weights of the RNN, 
similarly to the knowledge encoded in search trees in MCTS methods.

\section{Methods}
The RNN-based policy gradient setup described above is a powerful way to search combinatorial spaces \citep{bello2017neural}.
However, we show that this framework can suffer from an \textit{early commitment} phenomenon and from \textit{initialization bias} in symbolic spaces, both of which limit exploration.
To address these issues, we present two novel exploration techniques for neural-guided SO.

\subsection{Hierarchical entropy regularizer}\label{sect:HE}
Policy gradient methods typically include an 
entropy regularizer $\mathcal{R}_H$ in (\ref{eqn:update-rule}) that is proportional to the entropy at each step of autoregressive sampling
\citep{abolafia2018neural}.
This standard entropy regularizer is given by:
\begin{align}
\mathcal{R}_H (\tau) = \sum_{i=1}^{|\tau |}  H[ \policy (\cdot | \tau_{1:(i-1)}; \theta) ],
\label{eqn:entropy}
\end{align}
where 
$H[\policy(\cdot)] = - \sum_{x \in X} \policy(x) \log \policy(x)$ 
is the entropy.
This entropy regularizer promotes exploration and helps prevent the RNN from converging prematurely to a local optimum.
Notably, (\ref{eqn:entropy}) is simply a sum across time steps $i$; thus, all time steps contribute equally.

When optimizing discrete sequences (or navigating RL environments with deterministic transition dynamics), 
each sequence can be viewed as a path through a search tree, as depicted in Fig.~\ref{fig:expression_space}.
As learning progresses (i.e., the entropy around promising paths is reduced), 
trajectories $\{\tau^{(j)}\}_{j=1}^{M}$ sampled from the RNN tend to satisfy 
$$\frac{1}{M}\sum_{j=1}^{M}  
H[ \pi(\cdot
| \tau_{1:(i-1)}^{(j)}; \theta) ]
\leq
\frac{1}{M}\sum_{j=1}^{M}    
H[ \pi(\cdot
| \tau_{1:(k-1)}^{(j)}; \theta) ], \ \forall\ 0 < i < k,$$
by virtue of the sequential sampling process.
This causes the sum in (\ref{eqn:update-rule}) corresponding to the entropy regularizer (\ref{eqn:entropy}) to be concentrated in the later terms, while the earliest terms can quickly approach zero entropy.
When this happens, the RNN stops exploring early tokens entirely (and thus entire branches of the search space), which can greatly hinder performance.
This phenomenon, which we refer to as the ``early commitment problem,'' can be observed empirically, as shown in Figure \ref{fig:hierarchical_entropy} (top).
Note that the entropy of the first token of the sequence drops toward zero early on in training.

We propose a simple change to combat the early commitment problem: replacing the sum in (\ref{eqn:entropy}) with a weighted sum whose weights decay exponentially with a factor $\gamma < 1$.
Thus, our hierarchical entropy regularizer is given by:
\begin{align}
\mathcal{R}_H (\tau) = \sum_{i=1}^{|\tau|} \gamma^{i-1}  H[ \policy (\cdot | \tau_{1:(i-1)}; \theta) ].
\label{eqn:hierarchical_entropy}
\end{align}
Intuitively, using (\ref{eqn:hierarchical_entropy}) encourages the RNN to perpetually explore even the earliest tokens.

\subsection{Soft length prior}\label{sect:SLP}
Before training begins, using zero-weight initializers, the RNN emissions $\psi_{i}$ are zero.
Thus, the probability distribution over tokens is uniform at all time steps: $\policy(\tau_i) = \frac{1}{|\mathcal{L}|} \forall \tau_i\in\mathcal{L}$.
We can inform this starting distribution by including a \textit{prior}, i.e. by adding a logit vector $\psi_\circ \in \mathbb{R}^{|\mathcal{L}|}$ 
to each RNN emission $\psi_{i}$.
This transforms the likelihoods over tokens to $\policy(\cdot | \tau_{1:(i-1)}; \theta) = \textrm{softmax}(\psi_{i} + \psi_\circ)$.
For example, we can ensure that the prior probability of choosing a binary, unary, or terminal token is equal (regardless of the number of each type of token in the library) by solving for $\psi_\circ^{\textrm{eq}}$ below:
\begin{align*}
    \textrm{softmax}(\psi_\circ^{\textrm{eq}}) = \left(\frac{1}{3n_2}\right)_{n_2} \| \left(\frac{1}{3n_1}\right)_{n_1} \| \left(\frac{1}{3n_0}\right)_{n_0},
\end{align*}
where $(\cdot)_n$ denotes that element $(\cdot)$ is repeated $n$ times, $\|$ denotes vector concatenation, and $n_2, n_1, n_0$ are the number of binary, unary, and terminal tokens in $\mathcal{L}$, respectively.
The solution is:
\begin{align*}
    \psi_\circ^{\textrm{eq}} = (-\log n_2)_{n_2} \| (-\log n_1)_{n_1} \| (-\log n_0)_{n_0} + c,
\end{align*}
where $c$ is an arbitrary constant.

Under the prior $\psi_\circ^{\textrm{eq}}$, the distribution over expression lengths is heavily skewed toward longer expressions.
In particular, the expected expression length is $\mathbb{E}_{\tau\sim\psi_\circ^{\textrm{eq}}}[|\tau|] = \infty$ 
(see \cite{feller1957introduction} for an analogous discussion on random walks).
In practice, a \textit{length constraint} is applied; however, empirically this results in the vast majority of expressions having length equal to the maximum allowable length.
We show this empirically in Figure \ref{fig:length_distributions} (top).
This strong initialization bias makes it very difficult for the distribution to \textit{learn} an appropriate expression length.
To provide this capability, we introduce an additional \textit{soft length prior} to the RNN's $i$th emission:
\begin{align*}
    \psi_i^{\textrm{SLP}}= \left( \frac{-(i-\lambda)^2}{2\sigma^2} \mathbf{1}_{i>\lambda} \right)_{n_2} \| \left( 0 \right)_{n_1} \| \left( \frac{-(i-\lambda)^2}{2\sigma^2} \mathbf{1}_{i<\lambda} \right)_{n_0},
\end{align*}
where $\lambda$ and $\sigma$ are hyperparameters.
Note that the values of this prior depend on the positional index $i$ in the sequence.
In probability space, $\psi_i^{\textrm{SLP}}$ is a multiplicative Gaussian function applied to either binary tokens ($i>\lambda$) or terminal tokens ($i<\lambda$).
Thus, $\psi_i^{\textrm{SLP}}$ discourages expressions from being either too short ($i<\lambda$, where probabilities of terminal tokens are reduced) or too long ($i>\lambda$, where probabilities of binary tokens are reduced).

\section{Results and Discussion}

\begin{figure}[t]
\centering

\includegraphics[trim=0 0 0 0, clip, width=0.495\textwidth]{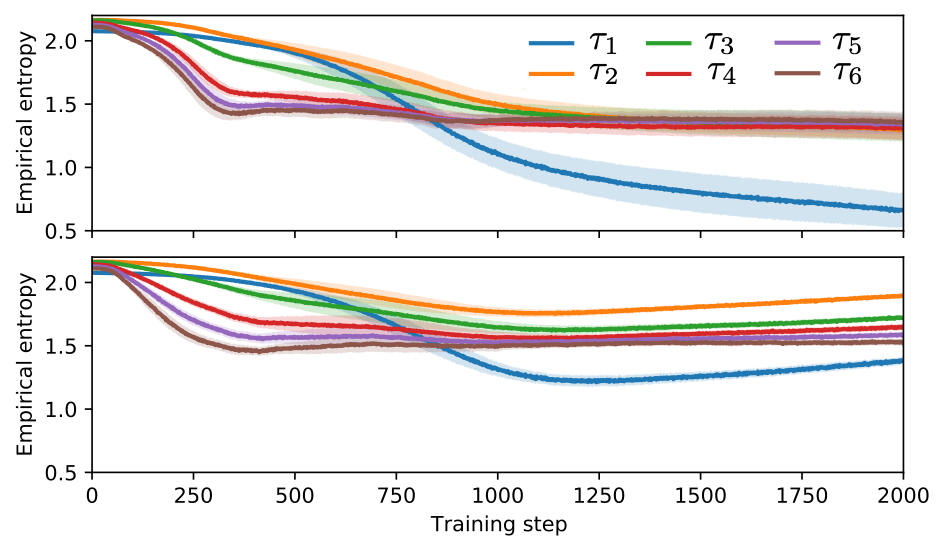}
\includegraphics[trim=0 0 0 0, clip, width=0.495\textwidth]{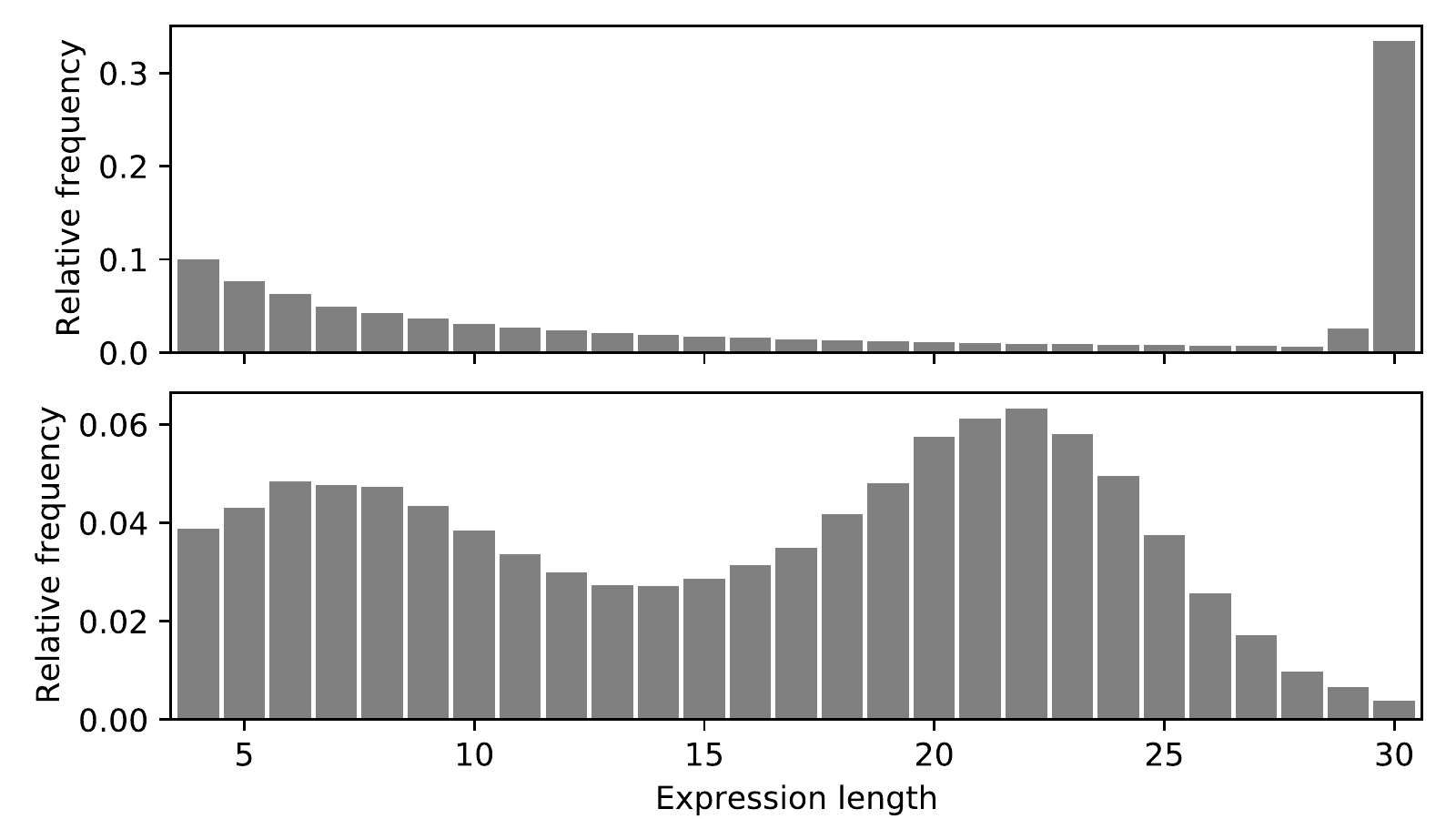}

\begin{minipage}[t]{.49\textwidth}
\centering
\captionof{figure}{
Empirical entropy of tokens $\tau_1$ to $\tau_6$ with standard (top) or
hierarchical (bottom) entropy regularizer on the Nguyen-7 SR benchmark.
}
\label{fig:hierarchical_entropy}
\end{minipage}
\hfill
\begin{minipage}[t]{.49\textwidth}
\centering
\captionof{figure}{
Initial distributions over expression lengths under the RNN without (top) or with (bottom) the soft length prior ($\lambda = 10, \sigma^2 = 20$).}
\label{fig:length_distributions}
\end{minipage}
\end{figure}

We demonstrate the benefits of our proposed exploration techniques on the standard Nguyen suite of SR benchmark problems proposed by \citet{uy2011semantically}.
The goal is to exactly recover benchmark symbolic expressions from a small synthetic dataset and limited set of tokens.
We build upon the open-source implementation of \textit{deep symbolic regression} given in \citet{petersen2019deep}.
We refer to this baseline approach as standard entropy (SE), then introduce the following additional variants:
hierarchical entropy (HE), 
soft length prior with SE (SLP; $\lambda=10, \sigma^2=5$),
and SLP with HE (SLP+HE).
For each variant, we perform a small hyperparameter grid search over 
$\eta \in \{0.001, 0.005, 0.01, 0.02, 0.03\}$ 
and
$\gamma \in \{0.7, 0.75, 0.8, 0.85, 0.9\}$.
All other hyperparameters are fixed, using values from \citet{petersen2019deep}.
We found best results with 
$\eta = 0.005$ for SE,
$\eta = 0.02$ and $\gamma = 0.85$ for HE,
$\eta =0.005$ for SLP, and
$\eta =0.03$ and $\gamma = 0.7$ for SLP+HE.

In Figure \ref{fig:hierarchical_entropy} (bottom), we demonstrate the effectiveness of the hierarchical entropy regularizer in ensuring that the RNN maintains diversity across time steps during the course of training, mitigating the tendency to choose paths with the same initial branches (i.e. $\tau_1$).
Figure \ref{fig:length_distributions} (bottom) shows that including the soft length prior results in a much smoother a priori distribution over expression lengths, reducing the bias toward long expressions.
In contrast to using a length constraint (i.e. \textit{hard} length prior) that \textit{forces} each expression to fall between a pre-specified minimum and maximum length, the soft length prior affords the RNN the ability to \textit{learn} the optimal length.

\setlength\intextsep{0pt}
\begin{wraptable}[9]{r}{7.5cm}
\caption{Average recovery rate, steps to solve, and length of best found expression across the 12 Nguyen SR benchmarks using SE, HE, and SLP.
}
\begin{tabular}{c|cccc}
Metric & SE & HE & SLP & SLP+HE  \\ \hline
Recovery & 83.8\% & 86.0\% & 84.0\% & \textbf{88.4\%} \\
Steps & 592.6 & 545.6 & 516.2 &  \textbf{429.1} \\
Length & 19.86 & 19.16 & 16.17 & \textbf{14.14} \\
\end{tabular}
\label{tab:sr-results-summary}
\end{wraptable}
In Table~\ref{tab:sr-results-summary}, we report recovery rate (fraction of runs in which the exact symbolic expression is found), average number of steps to solve the benchmark (up to 2,000), and the average length of the best found expression.
All metrics are averaged over 100 independent runs.
We observe improvements in each metric with both HE and SLP techniques separately: recovery rate improves, the search is more sample efficient, and the final expression is more compact.
The combination of both provides the best results.
Finally, it is worth noting that both contributions improve upon the state-of-the-art results for SR reported in \citet{petersen2019deep},
which already outperformed several commercial SR software solutions based on genetic algorithms and simulated annealing.

\section{Conclusion}
The combinatorial nature of symbolic spaces renders exploration a key component to the success of search strategies.
Here, we identified that existing policy gradient approaches for SO suffer from an early commitment phenomenon and from initialization bias, and we proposed two novel exploration techniques to address these limitations: a hierarchical entropy regularizer and a soft length prior.
We demonstrate that these techniques improve the state-of-the-art in neural-guided search for SR.

\subsubsection*{Acknowledgments}

This work was performed under the auspices of the U.S. Department of Energy by Lawrence Livermore National Laboratory under contract DE-AC52-07NA27344. Lawrence Livermore National Security, LLC. LLNL-CONF-820015.

\bibliography{iclr2021_conference}
\bibliographystyle{iclr2021_conference}

\end{document}